\documentclass[conference]{IEEEtran}
\IEEEoverridecommandlockouts
\usepackage{cite}
\usepackage{amsmath,amssymb,amsfonts}
\usepackage{algorithmic}
\usepackage{graphicx}
\usepackage{subcaption}
\usepackage{textcomp}
\usepackage{xcolor}
\usepackage{orcidlink}
\usepackage{comment}
\usepackage{multirow}
\usepackage{gensymb}
\usepackage{siunitx}
\usepackage{booktabs}
\usepackage{hyperref}

\hypersetup{
  colorlinks=false,
  urlcolor=black,
  linkcolor=black,
  linkbordercolor=white,
 urlbordercolor=white,
pdfborder={0 0 0}
} 

\def\BibTeX{{\rm B\kern-.05em{\sc i\kern-.025em b}\kern-.08em
    T\kern-.1667em\lower.7ex\hbox{E}\kern-.125emX}}
\begin{document}

\title{Maximum Temperature Prediction Using Remote Sensing Data Via Convolutional Neural Network\\
\thanks{This work was funded in part by the Horizon Europe project UP2030 (grant agreement n.101096405) and in part by the Project NODES through the MUR—M4C2 1.5 of PNRR  Grant ECS00000036.}
}

\author{
    \IEEEauthorblockN{
        Lorenzo Innocenti \orcidlink{0009-0009-6280-6230}, 
        Giacomo Blanco \orcidlink{0000-0001-6500-8511}, 
        Luca Barco \orcidlink{0000-0002-9089-9616}, 
        and
        Claudio Rossi \orcidlink{0000-0001-5038-3597}}
        \\
    \IEEEauthorblockA{
                \IEEEauthorrefmark{1}
                    LINKS Foundation\\ Turin, Italy \\
                    \textit{\{name\}.\{surname\}@linksfoundation.com}
    }
}

\maketitle

\begin{abstract}
Urban heat islands, defined as specific zones exhibiting substantially higher temperatures than their immediate environs, pose significant threats to environmental sustainability and public health. This study introduces a novel machine-learning model that amalgamates data from the Sentinel-3 satellite, meteorological predictions, and additional remote sensing inputs. The primary aim is to generate detailed spatiotemporal maps that forecast the peak temperatures within a 24-hour period in Turin. Experimental results validate the model’s proficiency in predicting temperature patterns, achieving a Mean Absolute Error (MAE) of 2.09°C for the year 2023 at a resolution of 20 meters per pixel, thereby enriching our knowledge of urban climatic behavior. This investigation enhances the understanding of urban microclimates, emphasizing the importance of cross-disciplinary data integration, and laying the groundwork for informed policy-making aimed at alleviating the negative impacts of extreme urban temperatures.
\end{abstract}


\section{Introduction}
Urban Heat Islands (UHIs) pose a complex problem in cities across Europe. These regions undergo increased temperatures as a result of urban development, human-induced heat sources, and shifts in climate. The implications of UHIs are extensive: they put pressure on energy networks, intensify health hazards during periods of intense heat, disturb ecosystems, and disproportionately affect disadvantaged communities. It is noteworthy that recent studies indicate a median rise of 45\% in the risk of death during severe heat incidents associated with UHIs \cite{huang2023economic}. Interdisciplinary approaches, encompassing green infrastructure, eco-friendly architectural design, and city planning, are necessary to alleviate UHIs. In 2023, Turin, similar to other urban hubs in Italy, demonstrated a significant number of days where temperatures exceeded the advised limits\cite{melis2023urban}.

This study concentrates on the creation of a machine-learning (ML) model that utilizes a Convolutional Neural Network (CNN) for maximum temperature prediction. This model is trained using the highest temperature readings recorded by weather stations, in conjunction with satellite data, meteorological and topographical information. The primary function of this model is to produce a high-resolution temperature prediction map for the city, offering a comprehensive forecast map for Turin's urban region.

\begin{figure}
    \centering
    \includegraphics[width=0.485\textwidth]{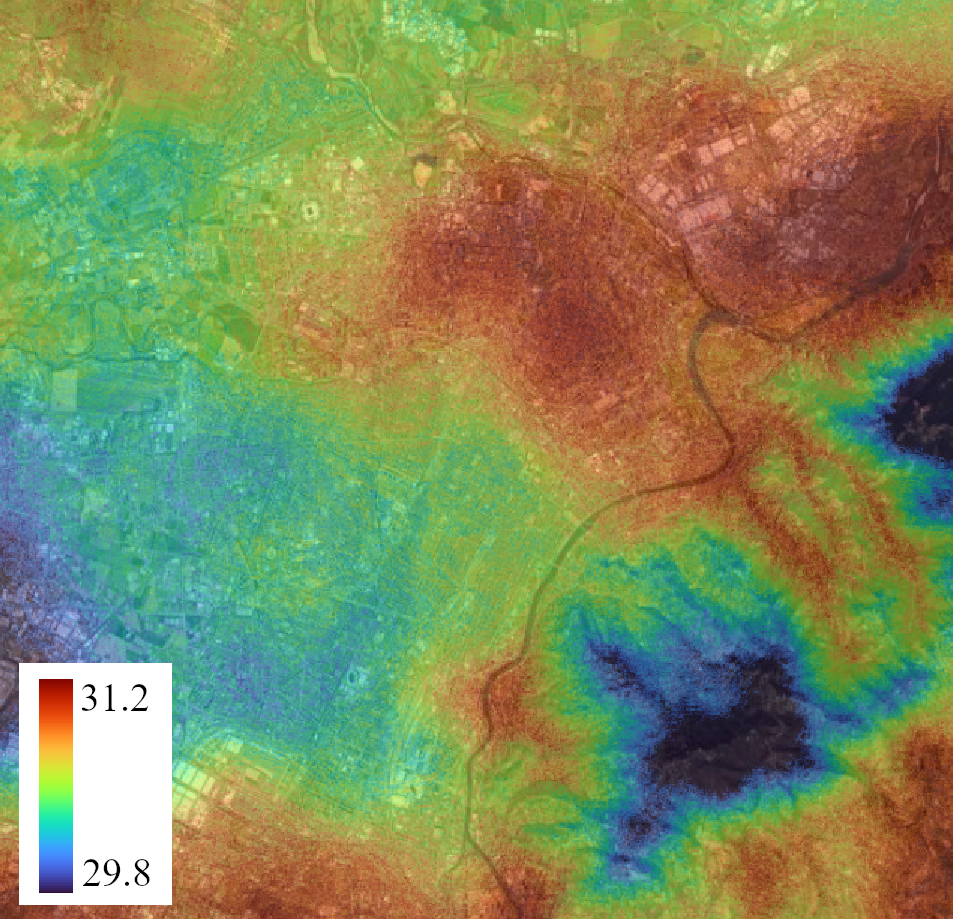}
    \caption{Output sample of the ConvNext model, for the day of Monday, June 12 2023, at a resolution of 50 m/px, superimposed on a satellite image of Turin. Legend displays maximum and minimum predicted temperatures, in Celsius degrees.} 
    \label{fig:heatmap}
\end{figure}

\section{Related works}
\label{sec:related}
The study of Urban Heat Islands (UHIs) and the prediction of air temperature ($T_a$) have been the subject of extensive research due to their substantial impact on urban sustainability and livability. Traditional approaches for predicting $T_a$ have largely relied on numerical weather prediction (NWP) \cite{Bauer2015}. However, these models encounter several challenges, primarily due to the computational expense and the difficulty in accurately capturing the variation of $T_a$ within urban regions.

The advent of cost-effective sensors has led to the establishment of dense urban climate monitoring networks in major cities. These networks offer an in-depth understanding of $T_a$ dynamics at the local scale, and the extensive databases they generate enable the application of data-driven methods for $T_a$ prediction. In contrast to numerical weather simulations, data-driven models can rapidly predict $T_a$ without requiring a comprehensive understanding of the related physical processes. When integrated with satellite-derived data and observations from various platforms, this approach proves effective for predicting $T_a$ at the urban level.

An early application of fully connected neural networks for temperature prediction is observed in the research by Gobakis et al. \cite{GOBAKIS2011104}, which focuses on UHIs prediction in Athens, Greece. The model is trained using historical temperature and solar radiation data to predict future temperatures. Notably, it offers predictions for specific locations where weather stations are located, rather than providing estimates for the entire city.
In more recent research, Oh et al. \cite{Oh2020}, akin to the previous study, develop a location-specific prediction ML model. Their model is notably trained using satellite-derived inputs, including albedo, land surface temperature, and land cover. This approach underscores the potential of data-driven deep learning models, particularly when combined with satellite data.
In a separate study, Lan et al. \cite{LAN2023101463} utilize fully connected neural networks to predict land surface temperature within the province of Changsha, China. They adopt a \textit{per pixel} approach, where the temperature prediction for each pixel is a function of the inputs specific to that pixel, disregarding relationships between pixels. The spatial resolution used is 1 km. In addition to the previously considered inputs, the study incorporates population density and elevation data. The model offers an annual estimation of UHI risk, indicating a coarse temporal resolution.

In conclusion, while numerous studies have explored ML techniques for UHI prediction, few have attained an urban level of resolution for estimating daily maximum temperatures—crucial for UHI assessment. This underscores the distinctiveness and importance of this study, which introduces a ML model employing a Convolutional Neural Network (CNN) for high-resolution spatiotemporal prediction of maximum temperatures throughout a day in Turin.
Furthermore, the presented approach goes beyond traditional point-based measurements. Instead, we provide densely estimated temperatures, offering a value for each pixel within the urban landscape. By combining high-resolution spatial predictions with dense temperature estimations, this study contributes to a more comprehensive understanding of UHI patterns and their impact on urban environments.

\section{Dataset}
\label{sec:dataset}
\subsection{Temperature Data}

The target for machine learning training is derived from temperature measurements, which are obtained from the Weather Underground temperature crowdsourcing platform\footnote{\url{https://www.wunderground.com}}. These stations provide temperature data at intervals that vary from 3 to 10 minutes, depending on the location. The present analysis considers a time period from 2018 to 2023. The maximum temperature recorded at each station each day is used as the ground truth for model training. 

\subsection{Satellite Data}

The satellite data used in this study is sourced from the European Space Agency's (ESA) Copernicus program, specifically from the Sentinel-3 satellite constellation. This constellation offers a spatial resolution of 1000 meters and a temporal resolution of two passages per day. The Sentinel-3 satellites are equipped with several sensors, including the Ocean and Land Color Instrument (OLCI), the Sea and Land Surface Temperature Radiometer (SLSTR), and the Synthetic-Aperture Radar Altimeter (SRAL). The machine learning model inputs include five specific Sentinel 3 bands, namely F1 and F2 (Thermal IR fire emission bands), and S7, S8, and S9 (Thermal IR Ambient bands). The goal of predicting the day's maximum temperature is achieved using only the morning passage data, collected between 9 AM and 11 AM. This data, presented as GeoTIFF files, encapsulates the daily value of the measured bands over the metropolitan city area of Turin.

In the dataset, numerous pixels are marked as non-valid in the data valid mask, indicating missing data. To address this, an imputation process is applied, which substitutes these missing values with estimated ones. This is only done when valid pixels constitute more than 75\% of the total, and if the condition is not satisfied, the data is not used. The chosen imputation method is bilinear interpolation, a technique that estimates the value of a pixel based on a weighted average of the nearest pixels. This approach maintains the integrity of the dataset, contributing to the robustness of the model's predictions.

\subsection{Weather Data}

The meteorological data used in this study is sourced from the Visual Crossing provider. This dataset encompasses fifteen distinct atmospheric indicators, each available at a daily resolution. Notably, these indicators lack spatially resolved information, yielding a single value for the entire analyzed area. The dataset exhibit no missing values, obviating the necessity for post-processing procedures.

\subsection{Topological Data}

Regarding topological data, two sources have been considered. The first one is the Digital Elevation Model (DEM), providing insights into the terrain’s morphology. The DEM, obtained as a GeoTIFF, represents a single constant observation throughout the reference period, as its main feature, the terrain’s altitude, remains static.

The second source of topological information is derived from the Copernicus Urban Atlas, which classifies land use in urban environments into 27 distinct classes. This dataset, updated to 2018, constitutes a singular observation for the entire reference period, even though land use information may change over time due to urban modifications. Similar to the DEM, this data is obtained as a GeoTIFF within the specified area of interest. To derive meaningful insights, the 27 Urban Atlas classes are initially grouped into 10 broader categories.

\section{Methodology}
\label{sec:method}
\subsection{Problem Formulation}

The model's input data is 3D matrix $I_t \in \mathbb{R}^{c \times n \times m}$, where $n$ and $m$ denote the image dimensions, and $c$ signifies the number of channels. The index $t$ indicates the day of the dataset. The dimensions $n$ and $m$ are determined by selecting a rectangular area that includes the city of Turin and its surroundings (bounding box), setting the resolution $r$, and deriving the pixel grid.
The input channels include five spectral channels from Sentinel-3 and an additional channel for DEM data. These channels are uniformly extracted using the same bounding box and resized to dimensions of $n \times m$. Land cover information is derived from the same bounding box, initially downsampled from its original high-resolution version, and then processed into a 10-channel matrix of size $n \times m$. Each channel corresponds to a unique land cover class, with its value indicating the percentage of that class's presence in the higher-resolution data. The weather component consists of fifteen channels, each containing the same repeated value for every pixel. Two additional channels encode the spatial coordinates of each pixel within the image. To incorporate temporal context, six extra channels use circular encoding for the day of the year, day of the week, and time of day of the Sentinel 3 passage. Notably, the value remains consistent across all pixels, resulting in values repeated throughout the entire image.
The dataset undergoes preprocessing steps to enhance model robustness, including normalization and geometric augmentations such as random flips, shifts, crops, and rotations.
The UHI detection task is defined as an image-to-image regression task, aiming to find a function $f_{\theta, t}: I_t \rightarrow \hat{T}_t$ where $\hat{T}_t \in \mathbb{R}^{n \times m}$ is a 2D image that represents the estimated maximum temperature for each pixel on the corresponding day $t$. During the training process, the output image $\hat{T}_t$ is compared to the ground truth $T_t \in \mathbb{R}^{n \times m}$, which is a matrix composed of pixels with the value equal to the average of the maximum temperature registered by each station which position falls in the area covered by the pixel for the day $t$. If no station is active in the area, the pixel is set to non valid.
For each valid pixel of $T_t$, the MSE loss is computed, and the value is used to update model weights $\theta$.

\subsection{Models Architecture}

In the field of temperature forecasting, two image-to-image architectures are suggested, inspired by the U-Net framework. Both comprise an encoder-decoder structure, with the encoder component extracting hierarchical features from the input data, while the decoder component reconstructs the spatial information. The models differ in the encoder's backbone, one being a ResNet-50 model and the other a ConvNext-tiny model.
Hierarchical features are extracted from input data by the encoder, while spatial information is preserved through skip connections. The decoder component consists of a cascade of upsample stages, comprising a residual block and an upsample layer in the case of the ResNet backbone and ConvNext-style upsampling block in the case of ConvNext backbone. Notably, features from the previous stage are concatenated with the skip connections, ensuring seamless information flow. To consolidate multi-channel feature maps into a single 2D representation, an additional block followed by a 1x1 convolutional layer is introduced. To mitigate artifacts caused by the convolutional model, the input matrix is mirror-padded to a fixed resolution.
In contrast to the aforementioned models, a shallow model is also proposed, which is a linear regression model. This model has a significantly lower computational complexity compared to the deep learning models. This makes it faster to train and infer, which can be an advantage in scenarios where computational resources or time are limited. However, the trade-off is that it may not perform as well in capturing complex patterns in the data.
The linear model takes the same input data as the previous models. However, unlike the U-Net inspired models, which can leverage the spatial relationships between pixels through their convolutional layers, it treats each pixel independently. This means that the model does not have the capacity to capture the spatial dependencies between different pixels in the image.

\section{Results}
\label{sec:results}
The performance of the model is evaluated using the mean absolute error (MAE) metric, which calculates the average absolute discrepancy between the predicted and actual temperature values. Three separate models are trained for each type, functioning at resolutions of 100, 50, and 20 meters per pixel. During the training phase, a batch size of 32 is used in conjunction with a learning rate of $10^{-3}$. A learning rate descent strategy is implemented, incorporating a patience of 20 epochs and a decay rate of 0.5. The AdamW optimizer \cite{loshchilov2017decoupled}, with a weight decay of 0.01, aids in regularization and optimization. The linear regression model is trained with SGD, using a learning rate of 0.1. The training process is conducted on a dataset covering the years 2018 to 2021, followed by model evaluation on data from the year 2023. The year 2022 is utilized as a validation set for early stopping and learning rate descent.

The results, displayed in Table \ref{tab:performance}, represent the average MAE for the year 2023. It is clear from the results that the linear regression model, despite its simplicity and reduced computational complexity, does not appear to be adequate for this type of prediction. The MAE values for the linear regression model are significantly higher than those for the ResNet and ConvNext models at all resolutions, indicating its struggle to capture the intricate patterns in the data. The ResNet model consistently outperforms, demonstrating lower MAE values across all resolutions. This finding suggests that the encoder-decoder architecture of the U-Net, which efficiently extracts hierarchical features while maintaining spatial information, is beneficial for temperature forecasting. However, the ConvNext model outperforms both the linear regression and ResNet models. Notably, despite its reduced complexity compared to ResNet, the ConvNext model achieves the lowest MAE values at all resolutions. Interestingly, there seems to be no correlation between the results and the resolution. This is counterintuitive, as one might expect the performance of the models to vary with the resolution. In conclusion, the image-to-image approaches, specifically the ConvNext model, yield satisfactory results for temperature forecasting. However, the specific requirements and constraints of the application, such as computational resources and inference speed, should be taken into account when selecting the most appropriate model.

In Figure \ref{fig:heatmap}, a sample output of the ConvNext model is shown, overlaid on a satellite image of Turin. The predictions of the model visually align with the urban layout of the city. Notably, the city center and industrial zones display higher temperatures, while the green areas and the river exhibit lower temperatures, as expected.
\begin{table}[t]
    \normalsize
    \centering
\begin{tabular}{@{}lccc@{}}
\toprule
             & Shallow     & ResNet & ConvNext      \\ \midrule
Weights      & \textbf{40} & 72.5 M & 37.4 M        \\
MAE at 100 m/px & 27,6        & 2,35   & \textbf{2,05} \\
MAE at 50 m/px  & 27,6        & 2.22   & \textbf{2,07} \\
MAE at 20 m/px  & 27,6        & 2.18   & \textbf{2.09} \\ \bottomrule
\end{tabular}
    \caption{Performance of the three models at different resolutions, in degrees Celsius.}
    \label{tab:performance}
\end{table}

\section{Conclusion and future works}
\label{sec:conclusion}
This research introduces a machine-learning model designed for forecasting high-resolution maximum daily temperatures, leveraging satellite data. The constraints of this small-scale case study are acknowledged, while recognizing the potential for enhancements.

Future endeavors could potentially refine the predictive model by integrating multistep predictions. This would entail utilizing data from several preceding days for predictions and simultaneously forecasting multiple upcoming days. This approach could offer a more holistic perspective of temporal trends and patterns.

Exploring the integration of supplementary input data presents another promising direction. For example, the incorporation of population density data could yield meaningful insights, as densely populated regions may display distinct patterns compared to sparsely populated areas. Furthermore, the Normalized Difference Vegetation Index (NDVI), a measure of the presence and health of green vegetation, could be a beneficial addition. The integration of NDVI data could aid in comprehending the influence of vegetation on the variable under study. These proposed enhancements could markedly augment the model's predictive capabilities and offer a more detailed understanding of the factors impacting the predictions.


\bibliographystyle{unsrt}
\bibliography{references}

\end{document}